\documentclass[runningheads]{llncs}
\usepackage[T1]{fontenc}
\usepackage{graphicx}
\usepackage{array, booktabs, multirow, makecell}
\usepackage{hyperref}
\usepackage{xcolor}
\usepackage{subcaption}

\begin{document}
\title{Using LLMs for Knowledge Component-level Correctness Labeling in Open-ended Coding Problems}
\titlerunning{LLM-based KC-level Correctness Labeling}
%
\author{Zhangqi Duan\inst{1} \and
Arnav Kankaria\inst{1} \and Dhruv Kartik\inst{1} \and
Andrew Lan\inst{1}}
\authorrunning{Z. Duan et al.}
\institute{University of Massachusetts Amherst}
\maketitle              
\begin{abstract}
Fine-grained skill representations, commonly referred to as knowledge components (KCs), are fundamental to many approaches in student modeling and learning analytics. However, KC-level correctness labels are rarely available in real-world datasets, especially for open-ended programming tasks where solutions typically involve multiple KCs simultaneously. Simply propagating problem-level correctness to all associated KCs obscures partial mastery and often leads to poorly fitted learning curves. To address this challenge, we propose an automated framework that leverages large language models (LLMs) to label KC-level correctness directly from student-written code. Our method  assesses whether each KC is correctly applied and further introduces a temporal context-aware Code-KC mapping mechanism to better align KCs with individual student code. We evaluate the resulting KC-level correctness labels in terms of learning curve fit and predictive performance using the power law of practice and the Additive Factors Model. Experimental results show that our framework leads to learning curves that are more consistent with cognitive theory and improves predictive performance, compared to baselines. Human evaluation further demonstrates substantial agreement between LLM and expert annotations.

\end{abstract}
\section{Introduction}
Knowledge components (KCs), defined as fine-grained units of skill or concept, play a central role in student modeling and learning analytics. Accurately modeling student mastery at the KC level enables a wide range of educational applications, including diagnostic feedback for instructors, personalized learning through content sequencing for students, and curriculum design. From a cognitive science perspective, a well designed set of KCs should follow the power law of practice \cite{snoddy1926learning}: student error rates on a KC should decrease as they attempt it more times. This principle underlies the study of learning curves, which provide an interpretable lens into how students acquire and consolidate skills over time. 

Defining what constitutes a correct/incorrect student attempt on a KC during problem solving is important. This correctness information is required for interpretable student modeling methods such as Performance Factors Analysis (PFA)~\cite{Pavlik_Koedinger}. However, in most real-world student response datasets, \emph{KC-level} correctness labels do not exist; only \emph{problem-level} correctness labels are available~\cite{choi2020ednetlargescalehierarchicaldataset,10.1145/2876034.2893409}. Therefore, in many existing works that study learning curves, they simply use problem-level correctness as KC correctness~\cite{10.1145/3754508.3754514,shi2024knowledge}. This setup can be problematic when a problem is associated with multiple KCs, resulting in poor curve fit \cite{rivers2016learning}. The reason is that problem-level correctness labels are coarse-grained: a student may have mastered some KCs while still struggling with others, a distinction that is lost under a single correctness label.

Moreover, in domains such as computer science education, obtaining reliable KC correctness labels remains a significant challenge. Common open-ended programming problems often require students to simultaneously apply multiple KCs when they construct their code. As a result, manually obtaining KC correctness labels can be labor-intensive, prone to bias and errors, and may not be scalable. Prior work~\cite{2024.EDM-long-papers.5,rivers2016learning} on automated approaches for KC-level correctness labeling leverage Abstract Syntax Tree (AST) to identify syntactic nodes, followed by labeling KC correctness based on whether a node appears in both the student-written code and the closest reference (correct) code. While this approach leads to good learning curves, it primarily captures surface-level, syntactic KCs and may overlook higher level, algorithmic KCs that are critical to solution correctness. Moreover, such KCs are less interpretable than those expressed in natural language. \cite{shi2024knowledge} studies KC attribution by leveraging problem-level correctness via transfer learning, noting that incorrect submissions may still demonstrate correct KCs, but evaluate on only three expert-defined programming KCs.

Recent advances in large language models (LLMs) have led to new opportunities for fine-grained analysis of student-written code. LLMs have demonstrated strong capabilities in code understanding, explanation, and error diagnosis. 
Although prior work~\cite{duan2025automated,moore2024automated,ozyurt2024automated} has explored using LLMs for automated KC generation and tagging, no existing study has systematically investigated their use for KC-level correctness labeling in open-ended programming problems. \\


\noindent\textbf{Contributions}
In this paper, we propose an LLM prompting-based KC-level correctness labeling framework for open-ended programming problems \footnote{
Our code and prompt can be found at: \url{https://github.com/umass-ml4ed/KC-Corretness-Labeling}}. To our knowledge, this work is the first work to explore the use of LLMs for KC-level correctness labeling in student-written code. Specifically, we employ a chain-of-thought prompting strategy with in-context examples, guiding the LLM to reason about the presence of each KC and the correctness of its application in a given student code, and ultimately produce a binary correctness label for each KC. We also design a temporal context-aware Code-KC mapping mechanism that aligns KCs directly with individual student code submissions. This mechanism enables more precise attribution of KCs to specific coding constructs, resulting in more accurate and fine-grained KC correctness labels. We evaluate the resulting KC labels by analyzing their fit to learning curves and additive factor model (AFM)~\cite{Cen_Koedinger_Junker_1970}. Results show that our proposed framework consistently yields KC-level mastery trajectories that more closely follow the power law of practice than baselines. We also show, via a human evaluation, that LLM-generated KC correctness labels have substantial agreement with human judgment. Together, these findings suggest that LLM-based KC correctness labeling provides a scalable and cognitively grounded alternative to existing heuristics, with potential to improve the quality and interpretability of learning analytics.

\begin{figure}[t]
  \centering
  \includegraphics[width=\linewidth]{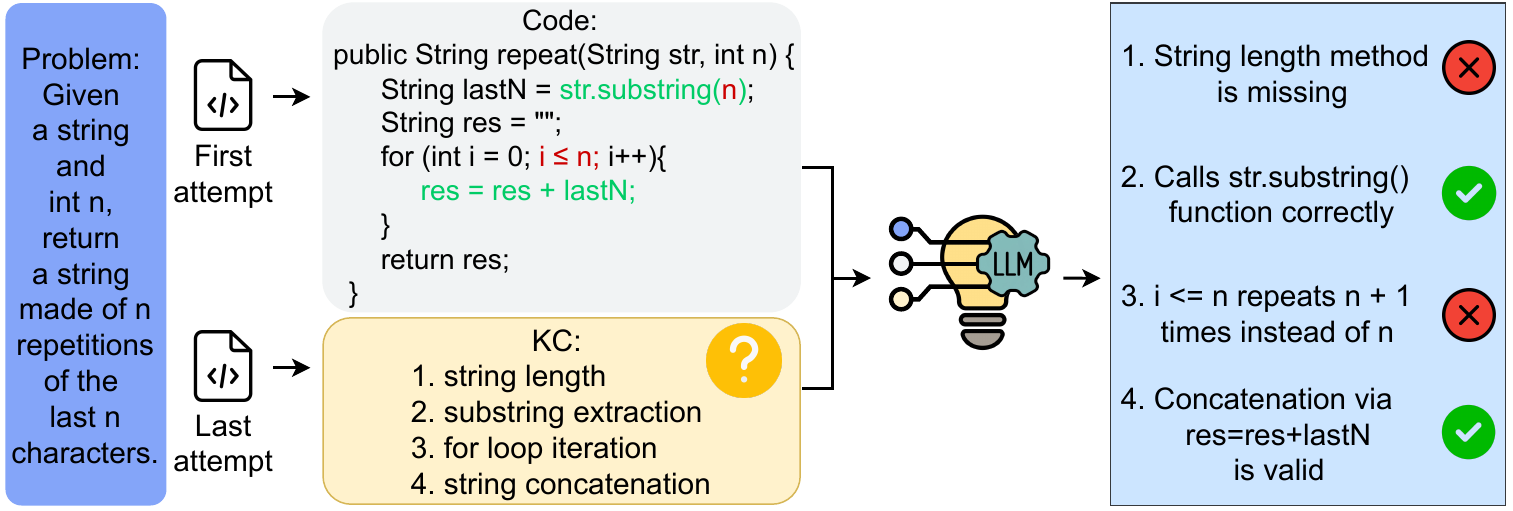}
  \caption{Illustration of our LLM-based KC-level correctness labeling pipeline for students' first attempt to open-ended coding problems. The set of KCs can either be given or selected based on the student’s last attempt. 
  }
  \label{fig:pipeline}
  \vspace{-.3cm}
\end{figure}

\section{Automated KC-level Correctness Labeling}
We now detail our automated LLM-based pipeline for labeling KC-level correctness in student code. We show a condensed example in Figure~\ref{fig:pipeline} for illustration. Our approach operates over two sets of different KCs. \\

\noindent\textbf{Prompting strategy}
For each problem–submission pair, we prompt GPT-4o~\cite{gpt-4o} in a chain-of-thought manner with three inputs: the problem statement, the student’s code submission, and the set of KCs mapped to the problem. We also include a few carefully constructed in-context examples, manually selected by the authors, as few-shot demonstrations in our prompt. These examples follow the exact same input-output format as the main prompt. We instruct the model to reason step-by-step: first determine whether a particular KC is used in the student code, and then assign a binary label indicating whether the student uses that KC correctly. If a KC required by a problem is not used in a student's code, it is labeled as incorrect. This structured prompting strategy encourages consistent and interpretable KC correctness judgments across student solutions. We perform labeling using greedy decoding with temperature=0 and top-p=1, yielding a single deterministic output for each code submission. \\

\noindent\textbf{Automated KC Generation}
Our framework applies to any KC sets. In most publicly available open-ended coding datasets \cite{codeworkout,10.1145/3545945.3569822,zhu2021programmingknowledgetracingcomprehensive}, each problem is tagged with KCs manually created by human domain experts. In contrast, recent work has explored automated, LLM-based KC generation and tagging, which we also experiment with in this work. We follow \cite{duan2025automated} and generate a set of KCs to account for the fact that many programming problems have multiple valid solution strategies. Human-created KC sets often overlook this variability; for example, in the CodeWorkout dataset \cite{codeworkout}, there are many KCs that are exercised fewer than three times across all problems, leading to poor learning curve fitting. Automated KC generation takes code diversity into account and alleviates this issue. By incorporating both KC sets, we aim to achieve broader and more faithful coverage of the conceptual space exercised by student solutions.
%
For each problem, we sample a diverse set of correct solutions by clustering CodeBERT~\cite{feng-etal-2020-codebert} embeddings and use these exemplars to prompt GPT-4o to generate candidate KCs. The generated KCs are then clustered and summarized to produce a refined KC set at a controlled level of abstraction. The procedure and hyperparameter settings follows \cite{duan2025automated}, which provides a complete description. \\

\noindent\textbf{Temporal Context-Aware Code-KC Mapping}
For LLM-generated KCs, we need to look further into code-KC pairs rather than problem-KC pairs for the generated KC mapping. Although the automated KC generation pipeline yields more comprehensive KC coverage for each problem, an individual student's code typically exercises only a subset of these KCs, i.e., those associated with the particular solution strategy it uses. To enable more accurate KC correctness labeling, we therefore adopt a context-aware mapping procedure to identify KCs for each student code.
First, for each representative correct code used in the KC generation pipeline, we prompt GPT-4o to select the subset of KCs used from the mapped KC set, and compute its CodeBERT embedding. Second, to associate each student code with the most relevant solution strategy, we collect both the first and last attempts of each student on this problem, and compute the CodeBERT embedding of the last attempt. Third, we identify the closest correct code using cosine distance since, empirically, the last attempt is a mostly-complete or near-correct version that includes all KCs the student uses in their solution strategy. In contrast, the first attempt often constitutes a highly incomplete draft version, such as a single return statement, which may not capture their intended solution strategy. Finally, we perform KC correctness labeling on the corresponding first attempt, use KCs associated with the correct code closest to the last attempt. Following prior work \cite{2024.EDM-long-papers.5,hoq2025patternbasedknowledgecomponentextraction,rivers2016learning,2023.EDM-long-papers.3}, we study learning curves with first attempts, since they correspond more to students' programming KC mastery compared to later attempts, which tend to involve additional debugging behaviors that we do not study in this work.

\section{Experiments}

\noindent\textbf{Dataset} We conduct experiments on the CodeWorkout~\cite{codeworkout} dataset, which was first used in the Second CSEDM Data Challenge~\cite{csedm}. The dataset contains $10{,}834$ codes written by $246$ students for $50$ open-ended programming problems, collected from an introductory Java programming course. We have access to both problem textual statements and human-created KC tags on each problem, among a total of $18$ KCs. For fair comparison with the human-created KCs, we set the number of generated KCs to $20$, yielding a comparable level of granularity. \\

\noindent\textbf{Methods and Evaluation Metrics} We experiment with three methods: 1) a frequently-used \textbf{Baseline} in the literature, where KC-level correctness of a code is simply set to the overall problem-level correctness (i.e., if a student code is incorrect, then all KCs are labeled as incorrect for this attempt); 2) our KC-level correctness labeling framework using \textbf{GPT-4o}, a strong, proprietary LLM; and 3) our framework using \textbf{Qwen3} Coder 30B A3B Instruct~\cite{qwen3technicalreport}, an open-weight LLM with strong coding and reasoning capabilities. Under each method, we also experiment with three different sets of KCs, to verify that our framework applies to any KC definitions: 1) \textbf{Human}-created KCs; 2) automatically-\textbf{Generated} KCs, and 3) KCs \textbf{Selected} by our temporal, context-aware code-KC mapping method. After getting the KC-level correctness labels using our method on each student code submission, we fit a learning curve under power law of practice and report the root mean square error (\textbf{RMSE}) and ${\bf r^2}$ metrics, both indicating how good the fit is. Since the power law of practice assumes a monotonically decreasing error curve, we restrict the curve fitting to have a non-positive exponential rate. In addition, following prior studies, we use the KC-level correctness labels to train a predictive Additive Factors Model (AFM), and report Area Under Curve (\textbf{AUC}) to measure its predictive performance on unseen students. \\


\begin{table*}[t]
\caption{Comparing performance on Learning Curve fit and AFM fit against correctness baseline over three KC sets.}
\centering
\scalebox{1}{
\begin{tabular}{p{0.13\linewidth}p{0.25\linewidth}>{\centering\arraybackslash}p{0.15\linewidth}>{\centering\arraybackslash}p{0.15\linewidth}|>{\centering\arraybackslash}p{0.13\linewidth}}
\toprule
\multirow{3}{*}{Method} & \multirow{3}{*}{KC Set} & \multicolumn{2}{c}{Power Law of Practice} &  AFM\\
\cmidrule{3-5}
& & RMSE $\downarrow$ & $r^2$ $\uparrow$ & AUC $\uparrow$\\
\midrule
\multirow{3}{*}{\shortstack{Baseline}}
 & Human & 0.110  &  0.253 & 0.529 \\
 & Generated & 0.109  &  0.269 & 0.532 \\
 & Selected & 0.083 & 0.320 & 0.538\\
\midrule
\multirow{3}{*}{\shortstack{GPT-4o}}
 & Human & 0.077 &  0.381 & 0.616 \\
 & Generated & 0.086 &  0.291 & 0.627 \\
 & Selected &0.069  & 0.383 & 0.631 \\
 \midrule
\multirow{3}{*}{\shortstack{Qwen3}}
 & Human & 0.079 & 0.386  & 0.665 \\
 & Generated & 0.070 & 0.417  & 0.676 \\
 & Selected & 0.082 & 0.362 & 0.629 \\
\midrule
\multirow{3}{*}{\shortstack{Qwen3 \\ w/o CoT}}
 & Human & 0.112 & 0.243  & 0.587 \\
 & Generated & 0.090 & 0.321  & 0.601 \\
 & Selected & 0.082 & 0.340 & 0.621 \\
\bottomrule
\end{tabular}}
\label{tab:learning_curve_result}
\vspace{-.3cm}
\end{table*}

\noindent\textbf{Quantitative Results} Table \ref{tab:learning_curve_result} shows the results of our experiments on learning curve and predictive modeling, averaged across all KCs within each KC set. We see that KC-level correctness labeling using both GPT-4o and Qwen3 significantly outperforms the ``problem-level-correctness as KC-level correctness'' baseline, on both RMSE and $r^2$, indicating a substantially better fit to the power law of practice. 
We also see the same trend on the future code correctness prediction task under AFM, although by a smaller margin in terms of AUC. These results show that fine-grained KC correctness labels not only improve learning curve fidelity, but also benefit with stronger predictive performance for student outcomes.
Comparing among different KC sets, we see that with any KC-level correctness labeling method, there is one set of automatically-generated KCs that outperforms human-created KCs: for KC-level correctness labeling via GPT-4o, the context-aware KC selection step leads to improved fit (e.g., $r^2$ of $0.383$ compared to $0.291$ without this selection step). However, when using Qwen3 for KC-level correctness labeling, the context-aware KC selection step results in lower performance, likely because Qwen3 is more sensitive to mismatches between the selected KCs and first attempt submission, as well as being more affected by instruction-following ambiguity. Nevertheless, both variants still consistently outperform the baseline, demonstrating that our KC-level correctness labeling framework is robust and generalizable across different KC sets. \\




\noindent\textbf{Ablation Study} We conduct an ablation study to examine the impact of the chain-of-thought (CoT) prompting strategy compared to a simpler prompting approach that directly outputs binary correctness labels using Qwen3. As shown in Table~\ref{tab:learning_curve_result}, incorporating CoT prompting, where the LLM performs explicit reasoning before generating labels, leads to a better fit to the power law of practice. This improvement is also reflected in the predictive performance under AFM, where the same trend is consistently observed. In contrast, compared to the baseline method, using a simple prompt without CoT yields only marginal improvements. These findings suggest that determining KC-level correctness is a non-trivial task that benefits from intermediate reasoning, as CoT prompting enables the model to better capture the underlying structure of student code and produce more accurate labels. \\

\noindent\textbf{Human Evaluation} To further validate our KC-level correctness labeling framework, we perform a human evaluation on the reliability of these labels by comparing them with human judgment. Two of the authors with extensive college-level Java coding experience conducted the same KC-level correctness labeling task independently, using the same instruction as the LLM, on $80$ randomly selected student-written codes, without knowing the LLM's labels. Results show that the Cohen's Kappa between two annotators is $0.86$, indicating almost perfect agreement; Cohen's Kappa score between humans and GPT-4o is $0.74$, indicating substantial agreement. This result further demonstrates the reliability of LLMs in accurately labeling KC-level correctness in student-written code. \\ 

\begin{figure}[t]
    \centering
    \begin{subfigure}[t]{0.48\columnwidth}
        \centering
        \includegraphics[width=\linewidth]{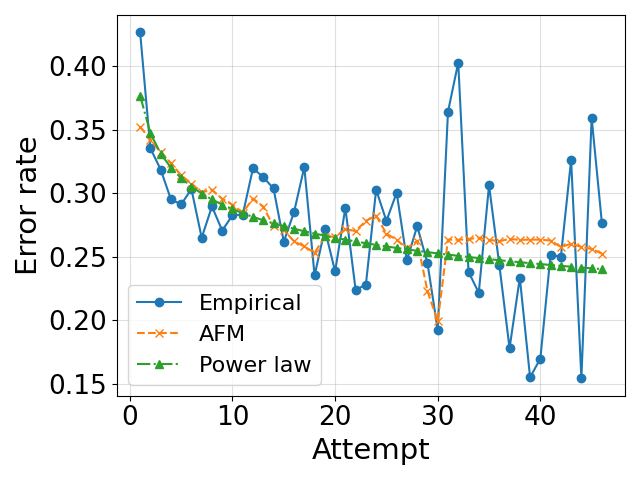}
        \label{fig:afm}
    \end{subfigure}
    \hfill
    \begin{subfigure}[t]{0.48\columnwidth}
        \centering
        \includegraphics[width=\linewidth]{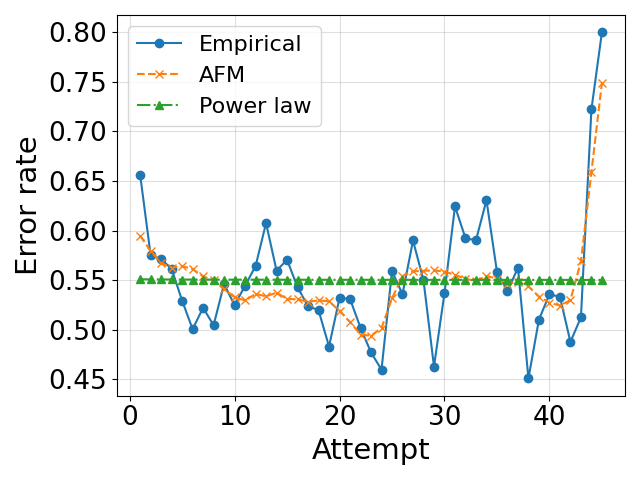}
        \label{fig:powerlaw}
    \end{subfigure}
        \vspace{-.7cm}
    \caption{Learning curves aggregated across all KCs, comparing LLM-generated KC-level correctness labels (left) with problem-level correctness labels (right).}
    \label{fig:learning_curve}
    \vspace{-.3cm}
\end{figure}

\noindent\textbf{Learning Curve Case Study}
We visualize several learning curves for KCs in the Selected KC set. We first fit learning curves for each KC independently and then average the error rates across KCs at each attempt across all students. Figure~\ref{fig:learning_curve} shows two sets of curves on error rate vs.\ attempt number, for empirical (observed) student correctness, the power law of practice fit, and AFM fit, comparing our KC-level correctness labeling method (left) and the problem-level correctness baseline. We see that using our LLM-based KC-level correctness labeling, the empirical error rate shows an overall decreasing trend despite noticeable fluctuations, reflecting heterogeneous problem difficulty. Both the power law of practice and AFM fitted curves closely follow this downward trajectory, indicating that the LLM-generated KC correctness labels are consistent with student behavior. In stark contrast, when using problem-level correctness as KC correctness, the empirical error rate fails to show a decreasing trend: it drops initially but then stops decreasing, before finally sharply increasing towards the end. As a result, fitting the power law of practice curve largely fails. This comparison highlights the benefit of using fine-grained, KC-level correctness in learning curve analysis. We also note that problem-level correctness labels have significantly higher empirical error rates than KC-level correctness labels. This observation further reinforces the need for the latter: even in incorrect codes, students can still show good mastery on a subset of the KCs required for the problem. \\


\section{Discussion and Conclusion}
In this work, we proposed a method to automatically label KC-level correctness for open-ended programming problems using LLMs. Our results show that replacing coarse problem-level correctness with fine-grained, LLM-generated KC-level correctness leads to learning curves that better fit the power law of practice and better predictive performance under AFM. This performance improvement also generalizes across different KC sets. Human evaluation shows substantial agreement between LLM-generated KC-level correctness labels and human judgment. Our framework also enables theory-grounded KC set evaluation. 
Beyond improvements in learning curve fit and predictive accuracy, accurate KC-level correctness labeling has direct implications for instructional decision-making in adaptive learning systems. Since KC mastery estimates drive feedback and content sequencing, coarse or incorrect labels can lead to misdiagnosis of student knowledge, resulting in inappropriate interventions (e.g., skipping needed practice or providing redundant support). In contrast, precise KC-level labels enable reliable detection of partial mastery and targeted feedback, ensuring that instructional decisions are aligned with students’ actual learning needs. \\

\noindent\textbf{Limitation and Future Work} Despite these promising results, several limitations remain. For example, the context-aware mapping assumes that final student code attempts accurately reflect their solution strategies, which may not always hold. Future work may incorporate factors such as problem difficulty into the labeling process. It would also be valuable to connect KC-level correctness labeling with test case correctness prediction, enabling more fine-grained evaluation of student code \cite{duan2024testcaseinformedknowledgetracing}. Furthermore, extending the framework to support student error simulation could provide deeper insights into common failure patterns and improve the robustness of automated feedback systems \cite{duan2026kaserknowledgealignedstudenterror}. Overall, our study suggests that LLM-based KC-level correctness labeling provides a scalable approach to KC-based learning analytics and student modeling.


%
%
%
\bibliographystyle{splncs04}
\bibliography{mybibliography}

\end{document}